\documentclass[a4paper, 10pt, conference]{ieeeconf}      

\IEEEoverridecommandlockouts                              

\usepackage{graphicx}
\usepackage[hidelinks]{hyperref}

\title{\LARGE \bf
SocialRobot: Towards a Personalized Elderly Care Mobile Robot 
}

\author{David Portugal$^{1}$, Lu\'{i}s Santos$^{1}$, Pedro Trindade$^{1}$, Christophoros Christophorou$^{1}$, \\Panayiotis Andreou$^{2}$, Dimosthenis Georgiadis$^{2}$, Marios Belk$^{2}$, Jo\~{a}o Freire$^{3}$, \\ Paulo Alvito$^{3}$, George Samaras$^{2}$, Eleni Christodoulou$^{1}$, and Jorge Dias$^{4}$
\thanks{$^{1}$David Portugal, Lu\'{i}s Santos, Pedro Trindade, Christophoros Christophorou and Eleni Christodoulou are with Citard Services Ltd., 1 Evrytanias Str., 2064 Strovolos, Nicosia, Cyprus. {\tt\footnotesize \{davidbsp, luissantos,pedrotrindade,christophoros,cseleni\} @citard-serv.com}.}%
\thanks{$^{2}$Panayiotis Andreou, Dimosthenis Georgiadis, Marios Belk and George Samaras are with the Department of Computer Science, University of Cyprus, P.O. Box 20537, 1678 Nicosia, Cyprus. {\tt\footnotesize \{panic,cspggd,belk,cssamara\}@cs.ucy.ac.cy}.}%
\thanks{$^{3}$Jo\~{a}o Freire and Paulo Alvito are with IDMind -- Engenharia de Sistemas, Lda., 1600-546 Lisboa, Portugal. {\tt\footnotesize \{jfreire,palvito\}@idmind.pt}.}%
\thanks{$^{4}$Jorge Dias is with the Institute of Systems and Robotics, University of Coimbra, 3030-290 Coimbra, Portugal, {\tt\footnotesize jorge@isr.uc.pt}, and with the Khalifa University of Science, Technology and Research (KUSTAR), Al Saada Street, Abu Dhabi 127788, UAE, {\tt\footnotesize jorge.dias@kustar.ac.ae}.}%
}

\begin{document}

\maketitle
\thispagestyle{empty}
\pagestyle{empty}

\begin{abstract}
SocialRobot is a collaborative European project, which focuses on providing a practical and interactive solution to improve the quality of life of elderly people. Having this in mind, a state of the art robotic mobile platform has been integrated with virtual social care technology to meet the elderly individual needs and requirements, following a human centered approach. In this short paper, we make an overview of SocialRobot, the developed architecture and the human-robot interactive scenarios being prepared and tested in the framework of the project for dissemination and exploitation purposes.
\end{abstract}

\section{INTRODUCTION}
\begin{figure}[t]
\centering
\includegraphics[width=0.65\columnwidth,keepaspectratio=true]{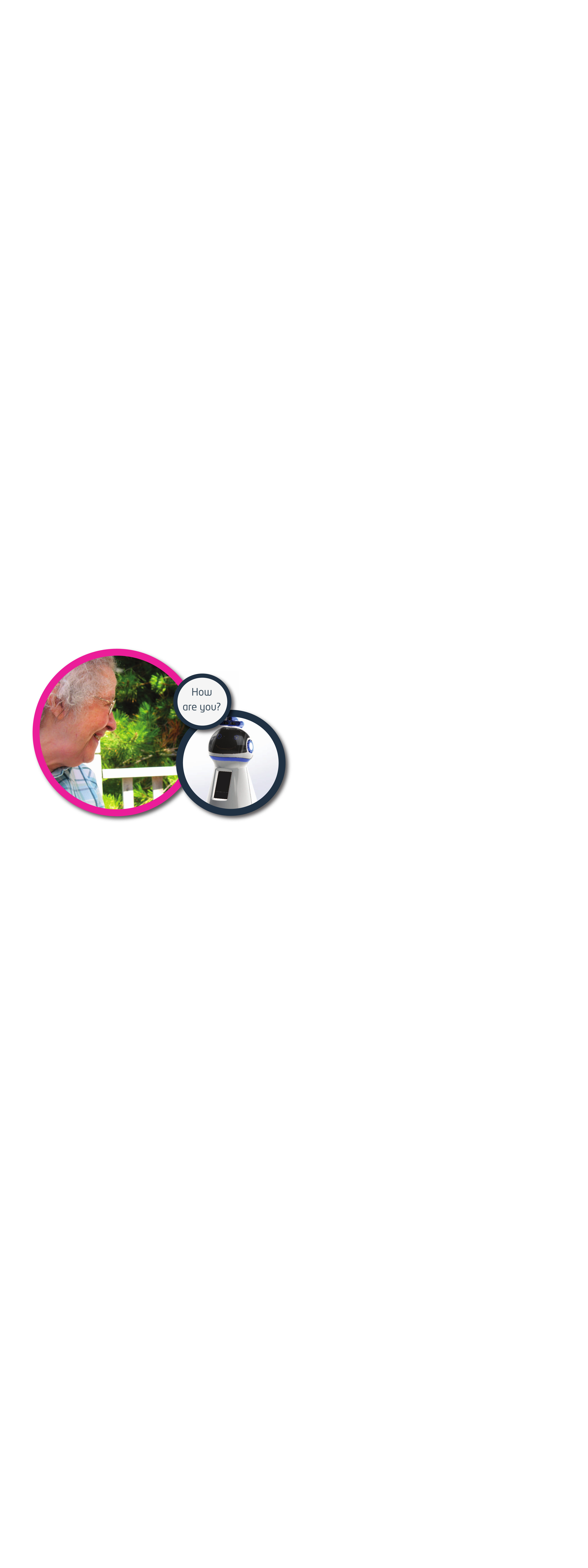}
\caption{Concept: An integrated Social Robotics system for ``Ageing Well".}
\label{fig1}
\end{figure}

Several demographic studies report that Europe's population is ageing, as the average life expectancy over the years increase~\cite{AAL}. Consequently, the elderly care market is growing, which in turn reveals a huge and unexplored potential. The SocialRobot Project~\cite{SR} aims to provide an answer to this demographic change challenge, through knowledge transfer and the creation of strategic synergies between the participating academia and industry partners. Therefore, an integrated Social Robotics system is being developed (\textit{cf.} Fig.~\ref{fig1}) to address key issues for improved independent living and quality of life of the elderly people.

The solution involves a practice-oriented elderly care mobile robot platform targeted to people with light physical or cognitive disabilities who can find pleasure and relief in getting help or stimulation to carry out their daily routine. The platform provides personalized services based on user information, their preferences and routines~\cite{pplidentification}, tackling initially the area of preventive care at an early stage of the ageing process. This is possible by integrating state of the art, standardized and interoperable robotic technologies and ICT-based care and wellness services, and benefiting from a virtual social care community network -- SoCoNet.

In the remaining of this paper, an overall description of the SocialRobot framework is provided, and a test scenario is presented. Lastly, conclusions are drawn upon the current state of the project.

\section{FRAMEWORK OVERVIEW}

In this project, a modular service robot architecture, following a user-driven philosophy, has been proposed. The social community model proposed encourages and supports communication, assistance and self-management of the elderly, promoting seamless connection and interaction to different people from all ages at any time, where the robot will act as a form of an intermediate agent between the elderly and the social care community. 

A Social Care Community Network (SoCoNet) was implemented so as to provide a secure web-based virtual collaborative social community network that enables the effective administration and coordination of the user profiles and virtual care teams (VCTs) around the elderly person. SoCoNet has been designed and maintained regardless of the robotic platform used, and it provides methods for retrieving and storing the required data for service provision. This way, it ensures a unique personalized profile of disabilities and abilities, special needs and preferences, stored in a secure database, thus promoting personalized care provision. Furthermore, SocoNet supports intelligent management techniques, which dynamically adapt the content included in the database throughout the elderly ageing process. These are statistical analysis techniques applied on the elderly daily monitoring information that enables the system to update preferences, priorities, routines and so on.

\begin{figure}[t]
\centering
\includegraphics[width=0.65\columnwidth,keepaspectratio=true]{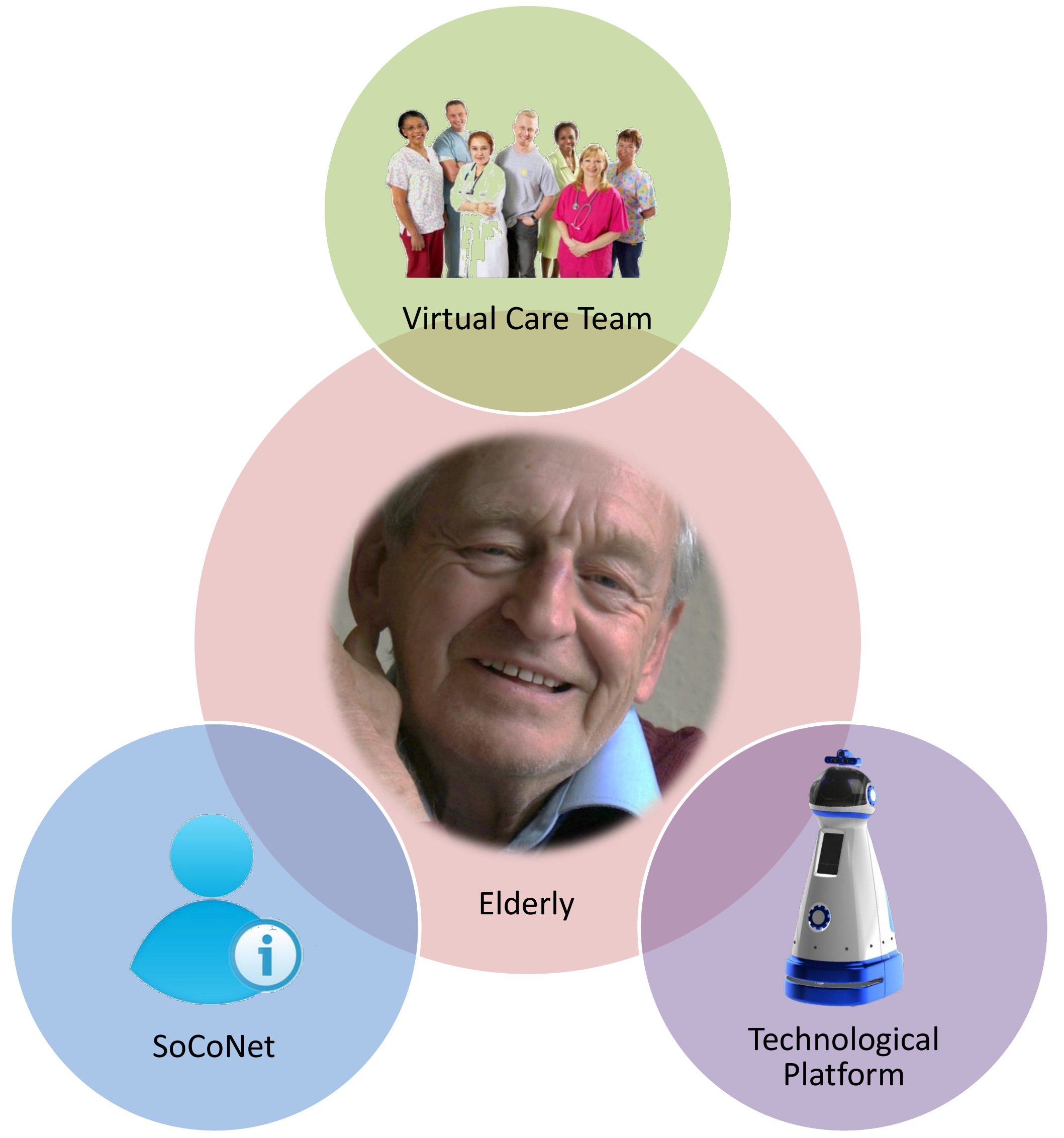}
\caption{SocialRobot Framework General Overview.}
\label{fig2}
\end{figure}

Services are actively provided by an appealing and affordable mobile robot platform~\cite{robot}, whose design considers the issues of size, shape, color and acoustic. The platform is a two wheels robotic base, with a structure body and robotic head with several integrated sensors such as an RGB-D sensor (Asus Xtion) and a laser range finder. This enables to fulfill the goal of promoting the maximum interaction between the elderly, family, friends, and carers supported by the robotic platform and the SoCoNet (\textit{cf.} Fig.~\ref{fig2}). On the technical side, the robot is fully integrated in the Robot Operating System (ROS), being capable of performing behavior analysis to adapt social relationships and contexts of the elderly people as they age, as well as mapping and navigating indoors in unstructured environments~\cite{SLAM} to provide affective and empathetic user-robotic interaction, taking into account the capabilities of and acceptance by elderly users. For more details on the intrinsic architecture underlying the SocialRobot system, please refer to~\cite{ehealth}.

\section{USE-CASE TEST SCENARIO}

After extensive testing of the different modules of the system, as well as their integration using ROS, a validation stage is currently underway. Since end-user involvement has been a priority ever since the beginning (\textit{e.g.} in the requirement specification stage), in this section a test scenario is described. This represents an exploitation activity scheduled to occur in the upcoming project review meeting. Involvement of the elderly in the age of 65 to 83 in the system design and prototype testing have shown positive end user acceptance related to the increase of motivation and reduction of hesitations in carrying out their daily routine with the support and company of the SocialRobot.

It is noteworthy that the system should provide ICT-based personalized services such as reminders and assistance, recognition of abnormal behavior and alerting, suggestions and guidance of daily activities, making use of innovative face recognition and vocal analysis. Having this in mind, the following scenario is envisioned:

\begin{enumerate}
\item Robot goes to a specific room (\texttt{\small{navigation}}).
\item Robot approaches a person (\texttt{\small{person tracking}}).
\item Robot recognizes person (\texttt{\small{face recognition}}).
\item Robot checks for medicine in a window of time (\texttt{\small{SoCoNet: check medicine}}).
\item Robot checks for activity in a window of time (\texttt{\small{SoCoNet: check activity}}).
\item Robot inquires the person (\texttt{\small{speak}}).
\item Robot extracts emotion from response (\texttt{\small{emotion recognition}}).
\item Robot suggests activity according to the emotional state (\texttt{\small{SoCoNet: suggest activity}}).
\item Robot resumes its previous task (\texttt{\small{navigation}}).
\end{enumerate}

The presentation of this preliminary scenario aims to attract both research and industrial stakeholders and promote know-how transfer in the project's technology and results at a European and an international level, so as to define a market penetration strategy.

\section{CONCLUSIONS}

In this work, an overview of the SocialRobot framework and a use-case test scenario has been presented. The ongoing project places emphasis in supporting the elderly to maintain their self-esteem in managing the daily routine, by addressing their security, privacy, safety and autonomy. The system not only considers the elderly as an active collaborative agent able to make personal choices, but also adapts the care model to his/her lifestyle, personalized needs and capabilities changes over the ageing process. Furthermore, it provides a platform that supports carers, both family members and therapists, in their daily tasks.

Innovation emerges from the human-robot interaction perspective (\textit{e.g.} emotion and face recognition, and empathetic interaction); the software perspective (\textit{e.g.} adaptation to the related context of daily routine occurrences as elderly age, and behavior modeling); the robotic perspective (\textit{e.g.} robot design, and navigation in unstructured environments); and the social care model perspective (\textit{e.g.} an elderly practice-oriented model integrating new types of social interaction, robotic monitoring and wellness services).





\section*{ACKNOWLEDGMENT}
This work is supported by the SocialRobot project, funded by the European Commission within the 7th Framework Programme FP7, by People Programme, Industry-Academia Partnerships and Pathways (IAPP), under grant agreement 285870.

\linespread{0.875}

\end{document}